\documentclass{article}

    \PassOptionsToPackage{numbers, compress}{natbib}

\usepackage{siunitx}
\usepackage[utf8]{inputenc} 
\usepackage[T1]{fontenc}    
\usepackage{hyperref}       
\usepackage{url}            
\usepackage{booktabs}       
\usepackage{amsfonts}       
\usepackage{nicefrac}       
\usepackage{microtype}      
\usepackage{xcolor}         
\usepackage{multirow}

\usepackage{graphicx} 
\usepackage{amsmath} 
\usepackage{fancyhdr} 
\usepackage{geometry} 
\usepackage{caption,subcaption}
\usepackage{tikz,graphics,float,epsf}
\usepackage{xcolor}
\usepackage{cancel}
\usepackage{wrapfig}

\usepackage{hyperref}
\hypersetup{colorlinks, citecolor=blue}

\usepackage[english]{babel}
\usepackage{csquotes}
\usepackage{enumitem}

\usepackage{amsfonts} 
\usepackage{natbib}
\usepackage{algorithm}
\usepackage{algpseudocode}
\usepackage{amssymb}
\usepackage{amsthm}
\usepackage{bbold}
\usepackage{amsmath}
\usepackage{mathtools}
\usepackage{xspace}

\theoremstyle{definition}

\theoremstyle{remark}

\theoremstyle{assumption}

\definecolor{es-blue}{rgb}{0,0.4,0.8}

\title{Maneuver Decision-Making Through Automatic Curriculum Reinforcement Learning Without Handcrafted Reward functions}

\makeatletter
\newcommand{\printfnsymbol}[1]{%
  \textsuperscript{\@fnsymbol{#1}}%
}

\author{Zhang Hong-Peng\thanks{\fontsize{8}{8} 
		Corresponding author.}\\
Aeronautics Engineering College, Air Force Engineering University
}

\begin{document}

\maketitle

\begin{abstract}
Maneuver decision-making is the core of unmanned combat aerial vehicle for autonomous air combat. To solve this problem, we propose an automatic curriculum reinforcement learning method, which enables agents to learn effective decisions in air combat from scratch. The range of initial states are used for distinguishing curricula of different difficulty levels, thereby maneuver decision is divided into a series of sub-tasks from easy to difficult, and test results are used to change sub-tasks. As sub-tasks change, agents gradually learn to complete a series of sub-tasks from easy to difficult, enabling them to make effective maneuvering decisions to cope with various states without the need to spend effort designing reward functions. The ablation studied show that the automatic curriculum learning proposed in this article is an essential component for training through reinforcement learning, namely, agents cannot complete effective decisions without curriculum learning. Simulation experiments show that, after training, agents are able to make effective decisions given different states, including tracking, attacking and escaping, which are both rational and interpretable.

\end{abstract}

\section{Introduction}
\label{sec:intro}

Autonomous maneuver decision-making is a sequential decision-making problem. In air combat, the goal of agents is to maneuver according to different states and launch missiles to defeat opponents. Usually, the agent is a human pilot. However, with the development of artificial intelligence, there have been many studies using programs or algorithms as virtual pilots for air combat maneuver decision-making in simulation environments, which aims to realize autonomous decision-making with unmanned combat aerial vehicles in future.

For example, Hu~\citep{hu2021application} studied autonomous maneuver decision-making on the basis of DQN~\citep{mnih2015human} for air combat maneuver decision-making, verifying the feasibility of deep reinforcement learning (RL) for air combat maneuver decision-making. Eloy et al.~\citep{garcia2021differential} applied game theory to the confrontation process of air combat and proposed a differential game method combined with the missile attack area in order to attack the static high-value targets. Dantas et al.~\citep{dantas2022supervised} compared different methods of evaluating the most effective moment for launching missiles during air combat. They found that supervised learning on the basis of simulated data can promote the flight quality in beyond-visual-range air combat and increase the likelihood of hitting the desired target. Fan et al.~\citep{fan2022air} used asynchronous advantage actor critic algorithm~\citep{mnih2016asynchronous} to address the problem of air combat maneuver decision-making. They proposed a two-layer reward mechanism, including internal rewards and sparse rewards. The simulation results indicated that the method can reduce the correlation between samples through asynchronous training. Wang et al.~\citep{yuan2022research} adopted deep deterministic policy gradient for beyond-visual-range air combat and validated the effectiveness of the method in simulations. Huang et al.~\citep{changqiang2018autonomous} applied Bayesian inference and moving horizon optimization to air combat maneuver decision-making. The method adjusted the weights of maneuver decision factors by Bayesian inference theory, and then computed the control quantities by moving horizon optimization. Ide et al.~\citep{pope2021hierarchical} proposed a hierarchical maximum-entropy reinforcement learning with reward shaping on the basis of expert knowledge. This approach achieved a 2nd place among eight competitors in the final DARPA’ s AlphaDogfight Trials event. 

Although these studies are valuable and lays foundation for replacing human pilots with computers in real environments, there are still some shortcomings. First, in order to simplify the complexity of decision-making, many studies use discrete action spaces, whereas real action spaces are continuous. Second, in these studies, a target hit by a missile is usually simplified as entering the missile attack zone. However, in the real world, even if the target enters the missile attack zone, it may not be hit. Finally, these studies use handcrafted reward functions, namely, the agent gets a non-zero reward signal at each time step. The disadvantage of handcrafted reward functions is that it may not be necessary and it costs lots of efforts and time. 

To address the above issues, this paper proposes an air combat maneuver decision-making method named automatic curriculum reinforcement learning (ACRL), without using any handcrafted reward functions. First, ACRL uses continuous action spaces and sparse rewards in maneuvering decision-making. Second, miss distance is used as a criterion for determining the results of air combat, instead of the missile attack zone. ACRL automatically generates curricula, and the agent is trained and tested in the curricula to learn how to maneuver effectively. Meanwhile, ACRL does not use any human data or handcrafted reward functions, but only uses the results of air combat as the reward function, that is, the reward for win is 1, the reward for loss is $-1$, and the reward for draw is 0. Finally, ACRL algorithm is verified by ablation studies, and the ability of decision-making of the trained agent is tested by simulation experiments.

\section{Related work}
\label{sec:pre}
Graves et al.~\citep{graves2017automated} introduced a method of automatically selecting curricula based on the increase rate of prediction accuracy and network complexity in order to improve the learning efficiency. Matiisen et al.~\citep{matiisen2019teacher} introduced teacher-student curriculum learning, in which the former selects sub-tasks for the latter and the latter attempts to learn these sub-tasks. Automatic curriculum learning methods also produced great performance in addition of decimal numbers~\citep{hochreiter1997long} and Minecraft~\citep{guss2019minerl}. A goal proposal module was introduced in~\citep{zhang2020automatic}. This method prefers goals that can decrease the certainty of the Q-function. The authors investigated the certainty through thirteen robotic tasks and five navigation tasks, and illustrated great performance of the method. 

Castells et al.~\citep{castells2020superloss} created a new method which automatically prioritizes samples with a little loss to efficiently fulfill the core mechanism of curriculum learning without changing the training procedure. Experimental results on several different computer vision tasks indicate great performance. Self-paced curriculum learning approach is introduced in~\citep{jiang2015self}, which takes into account both the prior knowledge known before training and the learning process. Stretcu et al.~\citep{stretcu2021coarse} proposed a new curriculum learning method that decomposes challenging tasks into intermediate target sequences for pre-training the mode. The results shown that the classification accuracy of the method on the data set has been improved by $7\%$. Sukhbaatar et al.~\citep{sukhbaatar2017intrinsic} proposed an automatic curriculum learning method based on asymmetric self-play. The method assigns two different minds to an agent: Alice and Bob. Alice aims to provide a task, and Bob aims to fulfill the task. The core is that Bob can comprehend the environment and fulfill the final task faster by self-play. 

Rane used curriculum learning to solve tasks with sparse rewards~\citep{rane2020learning}. The experimental results shown that curriculum learning can improve the performance of agents. Pascal et al.~\citep{klink2022curriculum} interpreted the curricula as a task distribution sequence interpolated between the auxiliary task distribution and the target task distribution, and framed the generation of a curriculum as a constrained optimal transport problem between task distributions. Wu et al.~\citep{wu2022robust} proposed bootstrapped opportunistic adversarial curriculum learning, which opportunistically skips forward in the curriculum if the model learned in the current phase is already robust. Huang et al.~\citep{huang2022curriculum} proposed GRADIENT, which formulates curriculum reinforcement learning as an optimal transport problem with a tailored distance metric between tasks.

\section{Method}
\label{sec:method}

\subsection{Models} 
The aircraft model is listed as follows~\citep{williams2007three}:
\begin{align}
\label{eq:1}
    \begin{cases}
    \dot{x}=v\cos\gamma\cos\phi\\
    \dot{y}=v\cos\gamma\sin\phi\\
    \dot{z}=v\sin\gamma\\
    \dot{v}=g(n_x-\sin\gamma)\\
    \dot{\gamma}=\frac{g}{v}(n_z\cos\mu-\cos\gamma)\\
    \dot{\psi}=\frac{g}{v\cos\gamma}n_z\sin\mu\\
    \end{cases}
\end{align}
where x, y, and z are three-dimensional coordinates of the aircraft. $\gamma$ and $\phi$ are pitch angle and yaw angle, respectively. v represents aircraft speed. g is gravitational acceleration. $\mu$, $n_x$, and $n_z$ are control signals. The missile model is~\citep{jie2019air}:
\begin{align}
\label{eq:2}
\begin{cases}
\dot{x}_m=v_m\cos\gamma_m\cos\phi_m\\
\dot{y}_m=v_m\cos\gamma_m\sin\phi_m\\
\dot{z}_m=v_m\sin\gamma_m\\
\dot{v}_m=\frac{(P_m-Q_m)g}{G_m}-g\sin\gamma_m\\
\dot{\phi}_m=\frac{n_{mc}g}{v_m\cos\gamma_m}\\
\dot{\gamma}_m=\frac{(n_{mh}-\cos\gamma_m)g}{v_m}\\
\end{cases}
\end{align}
where $x_m$, $y_m$, and $z_m$ are three-dimensional coordinates of the missile. $v_m$ represents missile speed, $\gamma_m$ and $\phi_m$ are pitch angle and yaw angle, respectively. $n_{mc}$ and $n_{mh}$ are control signals. $P_m$, $Q_m$ and $G_m$ are thrust, resistance and mass, respectively:
\begin{align}
\label{eq:3-5}
P_m&=
\begin{cases}
P_0 \quad t \leq t_w\\
0 \quad t > t_w
\end{cases}\\
Q_m&=\frac{1}{2}\rho v_m^{2}S_mC_{Dm}\\
G_m&=
\begin{cases}
G_0-G_tt \quad t \leq t_w\\
G_0-G_tt_w \quad t > t_w
\end{cases}
\end{align}
where $t_w = 12.0 s$, $\rho = 0.607$, $Sm = 0.0324$, $C_{Dm} = 0.9$. $P_0$ is the average thrust, $G_0$ is the initial mass, $G_t$ is the rate of flow of fuel. K is the guidance coefficient of proportional guidance law.
\begin{align}
\label{eq:6}
\begin{cases}
n_{mc}&=K\frac{v_m\cos\gamma_t}{g}[\dot{\beta}+\tan\epsilon\tan(\epsilon+\beta)\dot{\epsilon}]\\
n_{mh}&=\frac{v_mK\dot{\epsilon}}{g\cos(\epsilon+\beta)}\\
\beta&=\arctan(r_y/r_x)\\
\epsilon&=\arctan(r_z/\sqrt{(r_x^{2}+r_y^{2})})\\
\dot{\beta}&=(\dot{r}_yr_x-\dot{r}_xr_y)/(r_x^{2}+r_y^{2})\\
\dot{\epsilon}&=\frac{(r_x^{2}+r_y^{2})\dot{r}_z-r_z(\dot{r}_xr_x+\dot{r}_yr_y)}{R^{2}\sqrt{(r_x^{2}+r_y^{2})}}
\end{cases}
\end{align}
where $nmc$ and $nmh$ are the control commands of the missile. $\beta$ and $\epsilon$ are the yaw angle and pitch angle of line-of-sight, respectively. The line-of-sight vector is the distance vector $r$, where $r_x=x_t-x_m, r_y=y_t-y_m, r_z=z_t-z_m,$ and $R=\sqrt{(r_x^{2}+r_y^{2}+r_z^{2})}$. If the missile has not hit the target after 27 s, the target is regarded as missed. If the azimuth angle of the target relative to the aircraft exceeds $60^{\circ}$ (off-axis angle) at the same time as launching the missile, the target is regarded as missed. The conditions for the end of the simulation are: 1. One of the both side in air combat is hit by the missile (the missing distance is 30 m); 2. The missiles of the both sides miss their targets; 3. The simulation time has reached 100 s. Meanwhile, we do not use any handcrafted reward function. Namely, the reward obtained by the agent is 1 if it defeats the opponent, $-1$ if it is defeated by the opponent, and 0 in other cases.

\subsection{Proximal Policy Optimization} 
Policy gradient method is to calculate an estimate of policy gradient and use it for stochastic gradient ascent~\citep{sutton1999policy}. Both trust region policy optimization (TRPO)~\citep{schulman2015trust} and proximal policy optimization (PPO)~\citep{schulman2017proximal} are policy gradient methods. PPO is built on TRPO and it improves TRPO effectively. The objective function of TRPO is constrained by the policy update. After linear approximation of the objective and quadratic approximation of the constraint, the problem can be approximately solved by the conjugate gradient method. However, TRPO is relatively complex and incompatible with noise and parameter sharing. 

PPO is a better method which achieves the data efficiency and reliable performance of TRPO and it only uses first-order optimization. Schulman proposed a new target with a clipped probability ratio to generate an estimate of the lower bound of policy performance. The standard policy gradient method performs a update on each sample while PPO alternates between sampling data from the policy and performing several times of optimization on the sampled data. Schulman compared different surrogate targets, including not clipped target, clipped target and KL penalties (with fixed or variable coefficients), and verified the effectiveness of PPO.

Specifically, PPO modifies the objective function to penalize policy changes that cause the probability ratio to deviate from 1. The objective function of PPO is:
\begin{align}
\label{eq:7}
L^{CLIP}(\theta)&=E[\min{r_t(\theta)A_t, clip(r_t(\theta),1-\epsilon,1+\epsilon)A_t}]\\
P_{r}(\theta)&=\frac{\pi_{\theta}(a_t|s_t)}{\pi_{\theta_{old}}(a_t|s_t)}
\end{align}
By clipping the probability ratio and modifying the surrogate target, PPO can prevent excessive updates and make the optimization process more concise and robust than TRPO.
PPO is applied in self-play to train air combat agents. An action or maneuver is output by the policy and applied to the air combat environment. Then, a transition $(s_t, a_t, r_t)$ is generated and stored in experience pool. After that, transitions are sampled and used to train the policy to improve the action it output.

\subsection{Automatic Curriculum Reinforcement Learning}

The goal of RL is to explore the environment to maximize returns~\citep{badia2020agent57,team2023human,jin2022high}. Usually, the agent at the beginning makes decisions randomly because it has not yet been trained. The agent is trained by the RL algorithm with the samples it acquired, which can enable it to learn how to get more returns. Then, the trained agent can acquire samples with more returns by means of actions it generates, and repeating this process continuously, the initial agent that can only generate random actions can be transformed into an agent that can generate actions with more returns. On the other hand, human data can be used to pre-train agents to enable them to make rational decisions at the beginning~\citep{silver2016mastering,vinyals2019grandmaster}, rather than completely random decisions. After that, RL algorithms are used to train agents for better decisions.

The training process of random agents is more concise and the training cost is lower than that of pre-trained agents (collecting effective expert data requires plenty of time and efforts, and training an effective agent requires plenty of time and efforts as well). Due to the randomness, agents have a certain exploratory nature, that is, they may be able to find more rewards in the environment through random behaviors. On the other hand, this may also be a disadvantage, for example, in an environment with sparse rewards, it is difficult for the agent to obtain rewards only by random behaviors, so it is difficult for the agent to complete the given task.

\begin{figure}[]
	\centering
	\includegraphics[width=\textwidth]{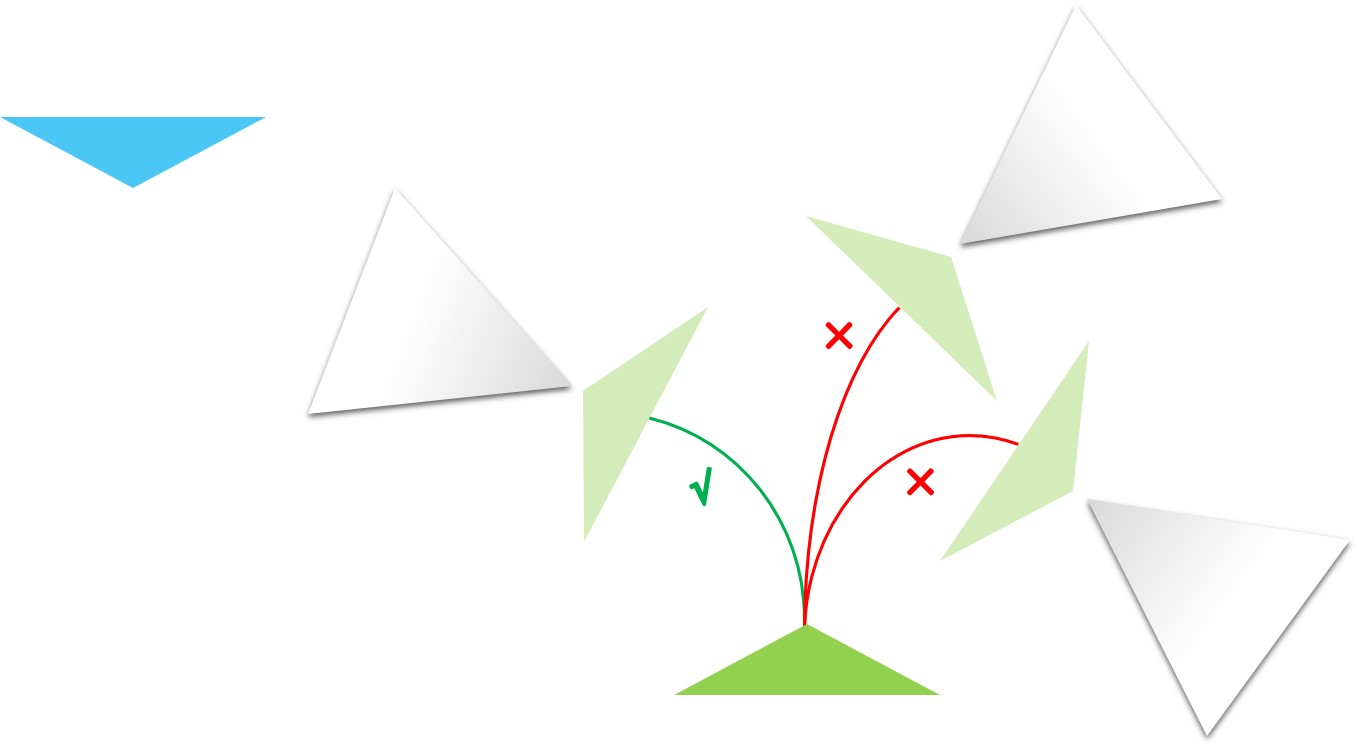}
	\caption{The schematic diagram of air combat maneuver decision-making.}
	\label{fig:abc}
\end{figure}

The maneuvering decision problem in air combat can be regarded as a task with sparse rewards. The schematic diagram of air combat maneuver decision-making is shown in Figure 2, which shows three possible results of corresponding decisions. The green aircraft and the blue aircraft represent the two sides of air combat. The solid lines represent the flight trajectories generated by the green aircraft, and the gray triangles represent the off-axis angle of the missile. As shown in the two red trajectories of Figure~\ref{fig:abc}, due to the fact that the azimuth angle of the target is greater than the off-axis angle, the missile will miss the target if it is launched. Only when the target azimuth angle is less than the off-axis angle, can the missile possibly hit the target, as shown in the green trajectory of Figure~\ref{fig:abc}.

It is difficult for agents to hit the target by random decisions, which increases the difficulty of training. In the process of training air combat agents by the RL algorithm, we found that they cannot make effective decisions. To solve this problem, we propose to use ACRL to train agents. ACRL decomposes the original task into a task sequence from easy to difficult, and then uses PPO to train the agent to complete all tasks in the task sequence, ultimately enabling the agent to complete the original task. In air combat, the initial state is randomly generated, such as the initial distance, initial velocity, and initial angle. When the initial azimuth angle of the opponent is larger, it is more difficult for the agent to overcome the opponent. Therefore, we propose a method for automatically generating a task sequence to improve the training efficiency. 

Concretely, perform several simulations with randomly generated initial states at first, with the initial angles and distances selected from $[-1, 1]$ (the original intervals are $[-180^{\circ}, 180^{\circ}]$ and $[4000 m, 16000 m]$, respectively. For simplicity, these two intervals are normalized to $[-1, 1]$), and record the initial angles and distances corresponding to the simulations in which missiles hit the targets. The maximum and minimum values of these initial angles and distances are $a_{\max}$, $a_{\min}$, $b_{\max}$, and $b_{\min}$, respectively. The four values form two intervals $[a_{\min}, a_{\max}]$ and $[b_{\min}, b_{\max}]$, which are proper subsets of $[-1, 1]$. Therefore, the initial angle and initial distance are first selected between these two intervals, which forms the first sub-task easier than the original task, that is, the agent is first trained in the two intervals. If the agent is able to make effective decisions in the two intervals, the interval length will be increased which corresponds to the second sub-task, $[a_{\min}-\delta, a_{\max}+\delta]$ and $[b_{\min}-\delta, b_{\max}+\delta]$. It is easier than the original task but more difficult than the first sub-task. After completing the second sub-task, the length of the interval is changed to $[a_{\min}-2\delta, a_{\max}+2\delta]$ and $[b_{\min}-2\delta, b_{\max}+2\delta]$, until the interval becomes $[-1, 1]$. is set to 0.1 and if the number of win is greater than 20 and the number of loss, the interval length will be increased. Otherwise, the interval will not be changed.

\subsection{Air Combat State}

As shown in Table~\ref{tab:state}, the input of the neural network is a one-dimensional vector with 11 elements : $\phi,\gamma,v,z,d,f_1,\phi_1,\gamma_1,d_1,\beta,f_2$. Min-max normalization is used and hyperbolic tangent function is used as the activation.

\begin{table}[]
	\centering
	\caption{Air combat state}
	\label{tab:state}
	\begin{tabular}{|c|c|}
		\hline
		\textbf{State}     & \textbf{Symbol}  \\ \hline
		yaw angle  &  $\phi$\\ \hline
		pitch angle                 & $\gamma$   \\ \hline
		velocity                 & $v$  \\ \hline
		altitude                 & $z$  \\ \hline
		distance between the two sides                 & $d$  \\ \hline
		launch missile            & $f_1$  \\ \hline
		yaw angle of the missile         & $\phi_1$  \\ \hline
		pitch angle of the missile    & $\gamma_1$  \\ \hline
		distance between the missile and the other side                & $d_1$  \\ \hline
		heading crossing angle     & $\beta$  \\ \hline
		launch missile from the other side   & $f_2$  \\ \hline
	\end{tabular}
\end{table}

\section{Experiments} 
\label{sec:exp}

The effectiveness of the proposed method is verified by ablation studies and simulation experiments in this section. The ablation studies compare the training process of ACRL and the original RL algorithm. Simulation experiments verify the decision-making ability of agents trained by ACRL. For each method, five independent experiments are conducted and the results are recorded, which include 40 iterations. 36 past agents are selected randomly in the test to fight against the current agent. The initial distance and initial angle are randomly selected within the corresponding intervals. Hyperparameters are shown in Table~\ref{tab:hyper}.

\begin{table}[]
	\centering
	\caption{Hyperparameters}
	\label{tab:hyper}
	\begin{tabular}{|c|c|}
		\hline
		\textbf{Name  }     & \textbf{Value}  \\ \hline
		Velocity  & $\left[250~\si{\meter\per\second},400~\si{\meter\per\second}\right]$   \\ \hline
		Batchsize                 & 1024   \\ \hline
		Optimizer                 & Adam  \\ \hline
		Actor learning rate                 & 0.002  \\ \hline
		Critic learning rate                 & 0.001  \\ \hline
		Actor architecture            & 256*256*4  \\ \hline
		Critic architecture         & 256*256*1  \\ \hline
		Activate function                 & tanh  \\ \hline
		Epoach                 & 6  \\ \hline
		$\gamma$                           & 0.99  \\ \hline
	\end{tabular}
\end{table}

\subsection{Ablation Studies}

\begin{figure}[]
	\vspace{-9ex}
	\centering
	\begin{subfigure}[b]{0.5\textwidth}
		\centering
		\includegraphics[width=\textwidth]{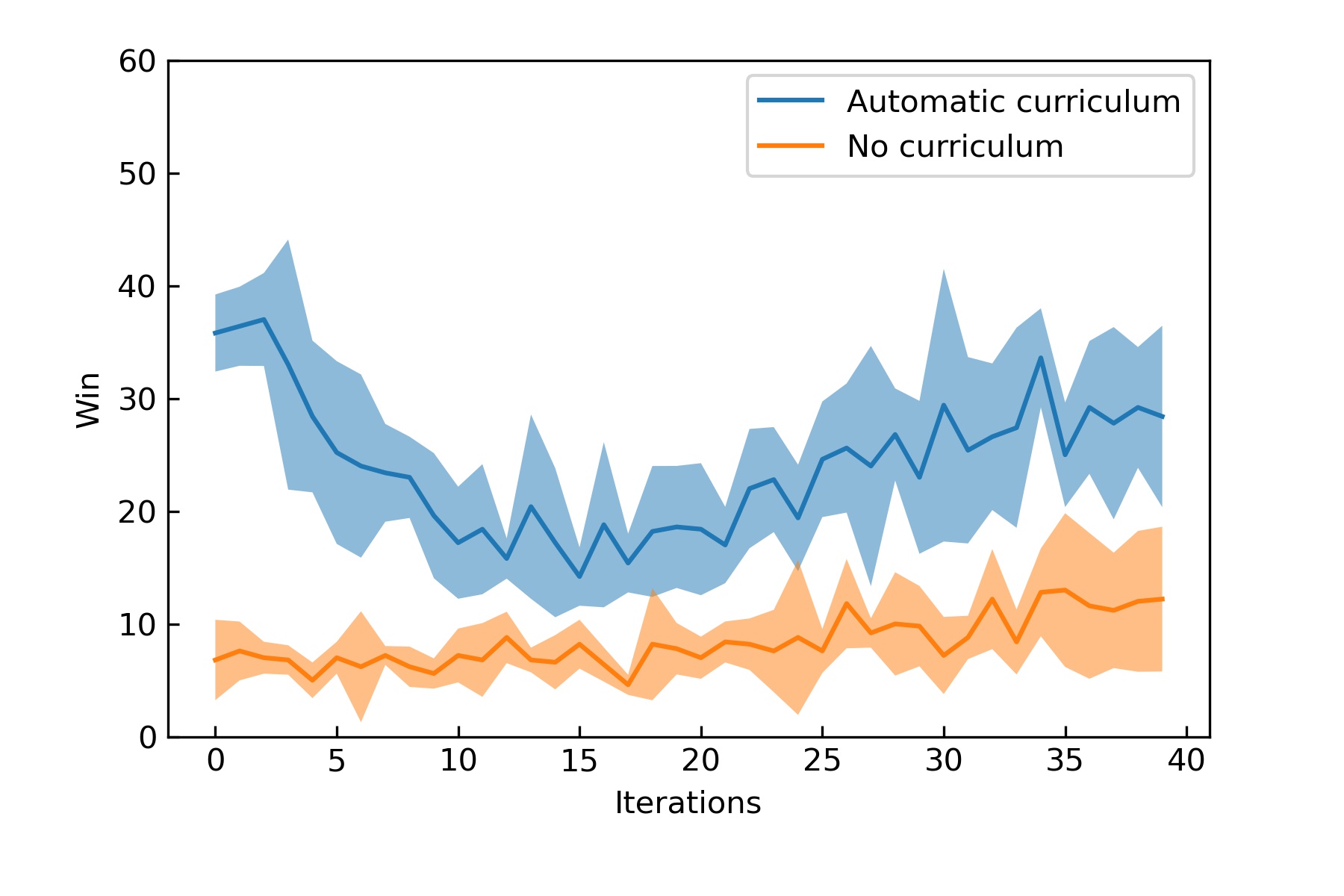}
		\caption{}
		\label{fig:win}
	\end{subfigure}
	\vfill
	\begin{subfigure}[b]{0.5\textwidth}
		\centering
		\includegraphics[width=\textwidth]{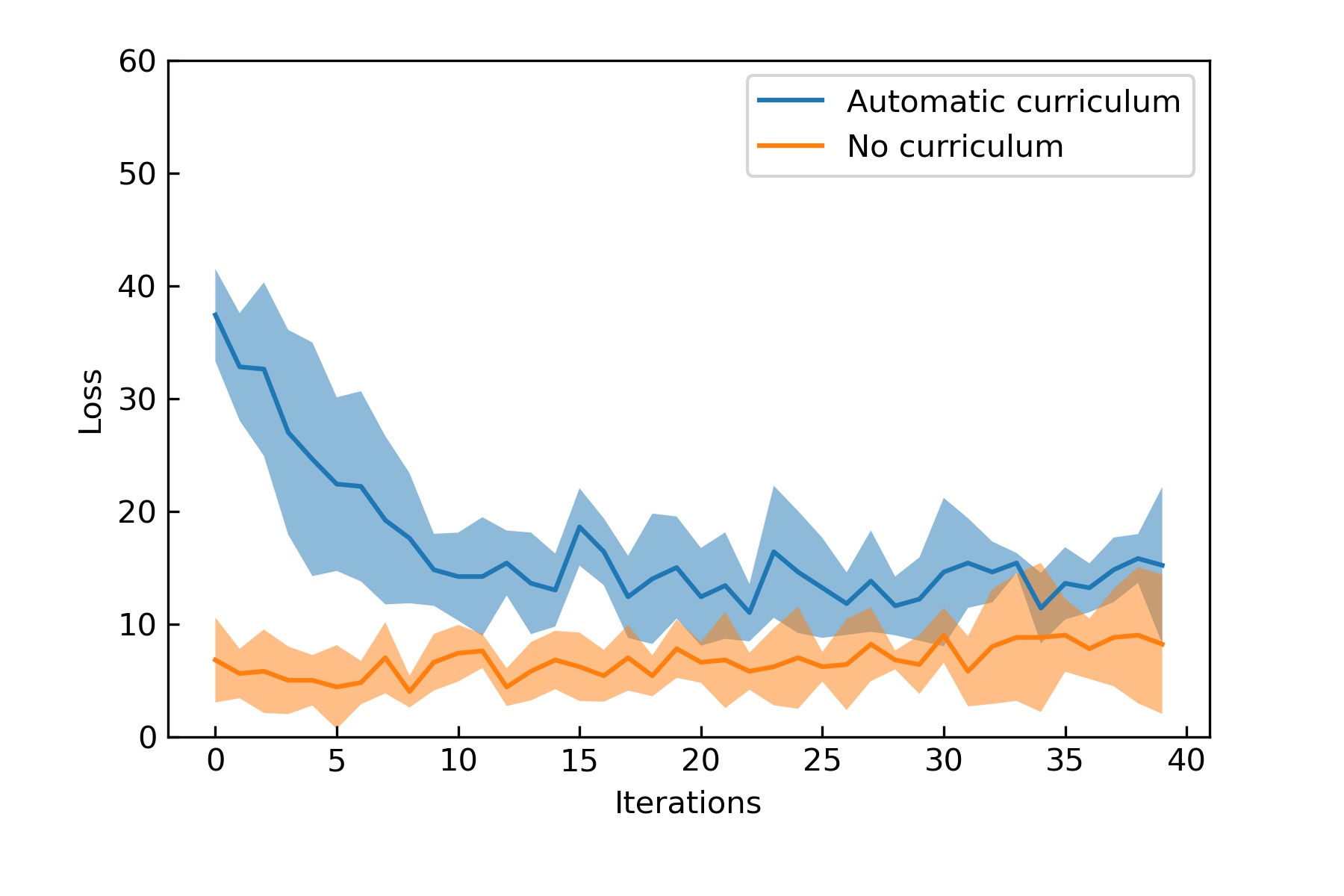}
		\caption{}
		\label{fig:loss}
	\end{subfigure}
	\vfill
	\begin{subfigure}[b]{0.5\textwidth}
		\centering
		\includegraphics[width=\textwidth]{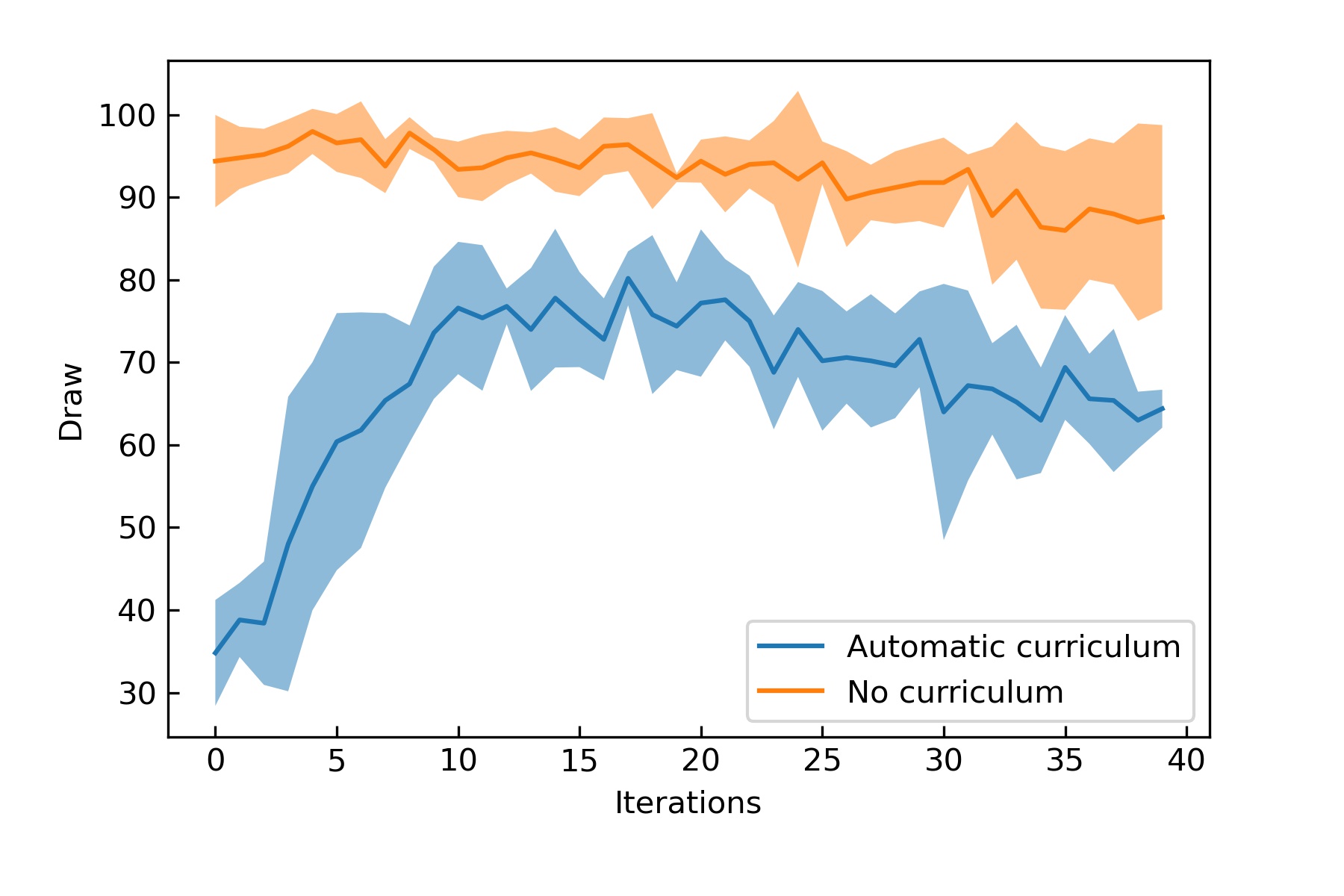}
		\caption{}
		\label{fig:draw}
	\end{subfigure}
	\vfill
	\caption{\textbf{(a)} Win. \textbf{(b)} Loss. \textbf{(c)} Draw.}
	\vspace{-4ex}
	\label{fig:train}
\end{figure}

Figure~\ref{fig:train} shows the changes in the number of win, loss, and draw in the training process of ACRL and the original RL algorithm. The solid line represents the mean of the number of win, loss or draw of the corresponding method, and the shaded part represents the standard deviation of the number of win, loss or draw.

As shown in Figure~\ref{fig:win}, in the training process, the number of win for ACRL is always greater than that for the original RL method. The reason why the number of win in ACRL first decreases and then increases is that the early curricula is easier, thus the missile is more likely to hit the target, resulting in more wins. As the difficulty of the curricula gradually increases, it becomes difficult for the missile to hit the target, so the number of win gradually decreases. At the same time, the agent gradually learns how to make effective decisions, resulting in a gradual increase in the number of win as shown in Figure~\ref{fig:win}. At the same time, more loss also represents better training, because both win and loss are the results of the agent's self-play.

From Figure~\ref{fig:draw}, it can be seen that the number of draw for ACRL is always less than that for the original RL method, indicating that agents trained by ACRL is more effective, as invalid decisions are more likely to lead to target missing which causes more draws. On the other hand, the number of draw for ACRL first increases and then decreases, which is consistent with the results in Figure~\ref{fig:win,loss}. This is because in early curricula, missiles are more likely to hit the target, resulting in fewer draws. As the difficulty of the curricula increases, it becomes more difficult to hit the target, resulting in more draws. The agent gradually learns effective decisions during training, which results in a decrease of the number of draw.

\section{Discussion}
\label{sec:Discussion}
In this article, agents use continuous action space, starting from completely random decisions, gradually learning how to make effective maneuver decisions through ACRL, and ultimately being able to cope with targets of different situations. On the other hand, previous researches discretized action space to reduce the complexity of maneuver decision-making, which is not practical since the real action space is continuous. Therefore, this method uses continuous action space for maneuver decision-making, and experimental results show that this method can handle continuous action space.

This method uses miss distance as the criterion for air combat, rather than whether the target has entered the missile attack zone. Previous researches use missile attack zones instead of miss distance, which is a simplified criterion. For example, when the target enters the missile attack zone, the distance between the missile and the target may still be thousands of meters, which is considered as a win in previous studies. In this article, the distance between the missile and the target needs to be less than 30 m to be considered as a win, which is more difficult than previous researches. Therefore, this study is more in line with the reality.

Designing reward functions is a time-consuming job, and unreasonable rewards can prevent intelligent agents from learning how to complete tasks. Previous researches usually uses handcrafted reward functions, such as angle reward functions and distance reward functions. ACRL does not require any handcrafted reward function, and only uses the results of air combat as reward signals. Meanwhile, according to Figure~\ref{fig:train}, it can be seen that ACRL can gradually increase the number of win during training, indicating that ACRL is a concise and efficient method. As shown in Figure~\ref{fig:train}, when using only RL, the number of win is always less, which demonstrates that the times of hitting the target is less, that is, the decisions made by agents are useless. These results indicate that the automatic curriculum learning proposed in this article is vital. Without automatic curriculum learning, agents cannot complete maneuver decision-making through RL.

\section{Conclusion}
\label{sec:Conclusion}
In this article, we propose ACRL to solve maneuver decision-making problems, which aims to use missiles to hit targets in different situations or avoid being hit by targets. ACRL has several advantages that previous methods do not have: Its action space is continuous rather than discrete, which makes it more realistic. It can make the missile hit the target instead of just aiming at the target. It does not need handcrafted reward functions, which makes it more concise and efficient. It can make the agent cope with targets in different situations, which indicates its effectiveness, because a target with larger azimuth angle is more difficult to defeat. The decisions made by agents are rational and interpretable. For example, agents learn to attack first and then move away to reduce the probability of being defeated by opponents without handcrafted reward functions.

\subsubsection*{Acknowledgments}
\vspace{-0.5em}
\small{
We thank Dr.Jung for his enlightenment.}

\normalsize
\bibliography{main.bib}

\end{document}